\documentclass[10pt, a4paper]{article}
\usepackage{lrec}
\usepackage[whole]{bxcjkjatype}
\usepackage{hyperref} 

\usepackage{soul}

\usepackage{url}

\usepackage{titlesec}
\titleformat{\section}{\normalfont\large\bf\center}{\thesection.}{1em}{}
\titleformat{\subsection}{\normalfont\SmallTitleFont\bf\raggedright}{\thesubsection.}{1em}{}
\titleformat{\subsubsection}{\normalfont\normalsize\bf\raggedright}{\thesubsubsection.}{1em}{}
\renewcommand\thesection{\arabic{section}}
\renewcommand\thesubsection{\thesection.\arabic{subsection}}
\renewcommand\thesubsubsection{\thesubsection.\arabic{subsubsection}}

\usepackage[utf8]{inputenc}
\usepackage{booktabs}
\usepackage{amsmath}
\usepackage{amsfonts}
\usepackage[pdftex]{graphicx}
\usepackage{mathtools}
\mathtoolsset{showonlyrefs=true}

\title{Video Caption Dataset for Describing Human Actions in Japanese}

\name{Yutaro Shigeto, Yuya Yoshikawa, Jiaqing Lin, Akikazu Takeuchi}

\address{STAIR Lab, Chiba Institute of Technology\\
         Tsudanuma 2-17-1, Narashino, Chiba, Japan\\
         \{shigeto, yoshikawa, lin, takeuchi\}@stair.center\\}

\abstract{
In recent years, automatic video caption generation has attracted considerable attention. 
This paper focuses on the generation of Japanese captions for describing human actions. 
While most currently available video caption datasets have been constructed for English, there is no equivalent Japanese dataset. 
To address this, we constructed a large-scale Japanese video caption dataset consisting of 79,822 videos and 399,233 captions. 
Each caption in our dataset describes a video in the form of ``who does what and where.''
To describe human actions, 
it is important to identify the details of a person, place, and action. 
Indeed, when we describe human actions, we usually mention the scene, person, and action. 
In our experiments, 
we evaluated two caption generation methods to obtain benchmark results. 
Further, we investigated whether those generation methods could specify ``who does what and where.''
\\ \newline }

\begin{document}

\maketitleabstract

\section{Introduction}
\label{sec:introduction}

\begin{figure*}[tb]
  \scriptsize
  \centering
  \includegraphics[scale=0.27]{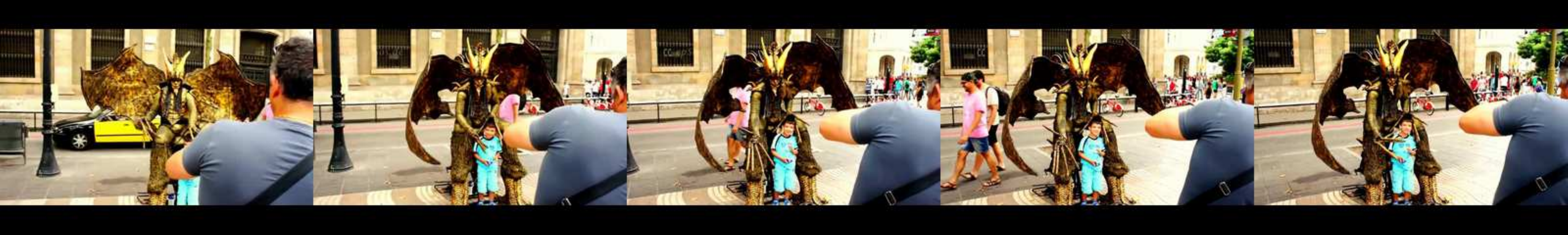} \\[5pt]
  \begin{tabular}{lll}
    \multicolumn{3}{c}{(a) Input video (1 fps).} \\[10pt]
    \toprule
    PLACE & PERSON & ACTION \\
    \midrule
    街中 & 青い洋服の男の子 & 写真を撮っている \\
    (the city) & (a boy with blue clothes) & (is being taken a photo) \\[5pt]
    屋外 & 青い服を着た男性 & 写真を撮っている\\
    (outdoors) & (a man worn blue clothes) & (is being taken a photo) \\[5pt]
    黒い柱のある道路 & 水色の服を着た少年 & 怪物のコスプレをした人と写真を撮ってもらっている\\
    (the road with black pillars) & (a boy worn light blue clothes) & (is being taken a photo with a person who made a cosplay of a monster) \\[5pt]
    車と黒い柱のある屋外 & 金色の仮装をした男性 & 立って子供を抱えている\\
    (outdoor space with car and black pillars) & (a man worn golden costumes) & (is standing and holding a child) \\[5pt]
    石造りの建物のある歩道 & 羽のついている金の衣装を着た人 & 子供と一緒に写真を撮っている\\
    (the pavement with a stone building) & (a person worn gold costumes with wings) & (is being taken a photo with children)\\
    \bottomrule
    \multicolumn{3}{c}{(b) Phrase annotations.}
  \end{tabular}

  \caption{
    An example of (a) an input video and (b) its phrase annotations.
    A sentence can be obtained by filling in the slots in the format:
    PLACE で PERSON が ACTION.
  }
  \label{fig:video-caption}
\end{figure*}

Automatic video caption generation is a task that outputs the description, or caption, of an input video~\cite{venugopalan2014translating,venugopalan2015sequence,yao2015describing}.
Video caption generation has many practical applications such as video searching using natural language queries, natural language video summarization, and use as a communication robot. 
It can also be useful for visually impaired people. 

This paper tackles one of such application, namely automatic human action description generation. 
To describe human actions, 
it is necessary to recognize and understand ``who does what and where.'' When we explain human actions, it is important to include details of the person, place, and action.
While various researchers have already introduced video caption datasets for describing human actions~\cite{Sigurdsson2016,Krishna2017a,2018trecvidawad}, 
none of these datasets evaluate those aspects individually.

Another problem is that there are difficulties in captioning resource-poor languages. 
Most previous research has focused on English caption generation due to the scarcity of resources targeting other languages. 
Each language is unique in terms of properties such as grammar and multi-word expressions, making it difficult to determine how to generate captions in other languages. 
Conversely, the practical applications of caption generation are common to all languages. 
Hence there is massive demand for caption generation in languages other than English. 

These issues motivated us to develop a video caption dataset for describing human actions in Japanese~(Sect.~\ref{sec:dataset}). 
This dataset is based on 79,822 videos collected from STAIR Actions, a dataset for human action recognition~\cite{stairactions2018}.
Each video has five descriptions on average, resulting in 399,233 captions in total. 
Each caption specifies ``who does what and where,'' 
and is written in Japanese.
This is the first instance of a Japanese video caption dataset, 
and is the most extensive dataset available, in relation to existing English caption datasets, although English and Japanese are clearly entirely different languages, and therefore, their statistics are not directly comparable.

In our experiments, 
we obtained benchmark results for this dataset~(Sect.~\ref{sec:experiment}), 
investigating whether captioning methods could specify ``who does what and where,'' in addition to standard generation evaluation such as BLEU, ROUGE, and CIDEr.

Our caption dataset is publicly available on our homepage.\footnote{{https://actions.stair.center}}

\section{Japanese Caption Annotations for STAIR Actions}
\label{sec:dataset}

\begin{table}[tb]
  \scriptsize
  \centering
  
  \begin{tabular}{l r r r r r r}
    \toprule
      & & &\multicolumn{4}{c}{The number of characters} \\
      \cmidrule(lr){4-7}
      & Uniq. & Voc.  & Mean & Median & Max  & Min  \\
    \midrule
    PLACE       & 49,460   & 5,214  & 6.3  & 6.0  & 60 & 1 \\
    PERSON      & 73,966   & 4,383  & 10.0 & 10.0 & 55 & 1 \\
    ACTION      & 110,926  & 10,098 & 11.9 & 11.0 & 73 & 1 \\
    \cmidrule(lr){0-6}
    sentence   & 306,116  & 13,836 & 30.2 & 28.0 & 135 & 8 \\
    \bottomrule
  \end{tabular}
  
    \caption{Our dataset statistics. 
  ``Uniq.'' indicates the number of unique phrases/sentences and ``Voc.'' is the vocabulary size.
  ``sentence'' (bottom row) represents statistics for sentences obtained using the template.
  }
  \label{tab:stats}
  
\end{table}

To construct the video caption dataset, we first collected videos from an existing video dataset and then asked workers to annotate multiple (approximately five) captions for each video, resulting in a dataset of 79,822 videos and 399,233 captions.\footnote{
We used an annotation service provided by BAOBAB Inc.} 

\subsection{Video Collection}

Videos were sourced from STAIR Actions dataset~\cite{stairactions2018}; a video dataset for human action recognition. 
Each video in this dataset contains a single human action from a set of 100 everyday actions (e.g., shaking hands, dancing, and reading a book). 
The average video length is approximately 5 seconds long, with a frame rate of 30 fps.
STAIR Actions dataset was thus a good fit with our objective of describing single actions (i.e., ``who does what and where''). 

\subsection{Caption Annotation}

Human actions can essentially be described in terms of ``who does what and where,'' with action descriptions typically mentioning the scene, person, and the specific action. 
On this basis, three elements were set as a requirement of our captions. 

To annotate the three elements, a question answering annotation procedure was performed. 
First, we asked workers the following questions about a video:
\begin{itemize}
  \setlength{\itemsep}{1.2pt}
  \setlength{\parskip}{0pt}
  \item Who is present? (PERSON)
  \item Where are they? (PLACE)
  \item What are they doing? (ACTION)
\end{itemize}
In this procedure, 
acceptable answers were a noun phrase for PERSON and PLACE and a verb phrase for ACTION. 

Further, we set the following annotation guidelines:
\begin{itemize}
  \setlength{\itemsep}{1.2pt}
  \setlength{\parskip}{0pt}
\item[(1)] A phrase must describe only what is happening in a video and the things displayed therein. 
\item[(2)] A phrase must not include one's emotions or opinions about the video. 
\item[(3)] If one does not know the location, write 部屋 (\textit{room}), 屋内 (\textit{indoor}), or 屋外 (\textit{outdoor}).
\item[(4)] If one does not know who the person is, write 人 (\textit{person}).
\end{itemize}
The phrases obtained were reviewed, and corrected if inaccurate. 
The annotation work was completed by 125 workers in four months. 
Figure~\ref{fig:video-caption} shows an example of our captions.

After phrase annotations were completed, 
sentences were obtained by complementing Japanese particles で and が: %
\begin{quote}
  PLACE で PERSON が ACTION.
\end{quote}

Obviously, this template-based sentence construction does not produce grammatically differing sentences. 
Since the objective of this research is to specify human actions, the captions may not require complex sentence patterns such as anastrophe and taigendome (a rhetorical device in Japanese); i.e., ending a sentence with a noun. 
Moreover, the sentences produced were not unnatural.

As a result, we obtained a total of 399,233 sentences. 
Table~\ref{tab:stats} shows the statistics for the annotated phrases and the sentences obtained using the template. 
As the table shows, the unique sentences account for 76.7\% of the total. 
For determining vocabulary size, 
we used KyTea\footnote{{http://www.phontron.com/kytea/}}~\cite{neubig2011pointwise}, a morphological analyzer, to tokenize each phrase/sentence into words.
In PLACE, the frequency of the terms (部屋, 屋内, and 屋外) was 118,092; 
i.e., these terms comprise one third of the phrases. 
There were 27,835 instances of 人; that is, 7\% of PERSON phrases.

\section{Related Work}
\label{sec:related-work}

\begin{table}[tb]
  \small
  \centering
  
  \begin{tabular}{lrr}
    \toprule
    Dataset                                    & \#videos      & \#captions    \\ 
    \midrule
    MSVD~\cite{Chen2010}                       & 2k           & 86k       \\
    TACoS ML~\cite{101007}               & 14k       & 53k    \\
    MSR-VTT~\cite{Xu2016}                      & 10k          & 200k      \\
    Charades~\cite{Sigurdsson2016}             & 10k          & 16k     \\
    LSMDC~\cite{Rohrbach2017}                 & 118k         & 118k      \\
    ActivityNet~\cite{Krishna2017a}            & 100k         & 100k      \\ 
    YouCook II~\cite{ZhXuCoCVPR18}                 & 15k & 15k \\
    VideoStory~\cite{D18-1117}                 & 123k         & 123k \\
    TRECVID~\cite{2018trecvidawad}          & 2k         & 10k \\
    \midrule
    Ours                                       & 80k       & 399k   \\
    \bottomrule
  \end{tabular}

  \caption{Video caption datasets. 
  }
  \label{tab:dataset}

\end{table}

Many video caption datasets have been constructed recently, including MSVD~\cite{Chen2010}, Charades~\cite{Sigurdsson2016}, ActivityNet~\cite{Krishna2017a}, and TRECVID~\cite{2018trecvidawad}. 
Table~\ref{tab:dataset} summarizes the video caption datasets most commonly used in video captioning experiments. 
Apart from MSVD, these datasets only provide English descriptions, 
and while MSVD contains 15 languages captions besides English, it has a limited number of captions (i.e., 6,245 captions at most) in other languages and none in Japanese language. 
Differing from MSVD, the dataset described in this paper provides descriptions in Japanese not English.

There are some existing video caption datasets for describing human actions. 
ActivityNet is a video caption dataset whose main objective is to detect and describe numerous events (human actions) in a long video (180 seconds on average), requiring the ability to recognize the dependencies between human actions. 
Conversely, each video in our dataset only contains just one action, 
which is appropriate for our research objective. 
Charades also provides descriptions of human actions. 
However, participant details are insufficient, with ``a person'' appearing frequently in the captions. 
Each sentence in TRECVID includes four elements of the video: Who, what, where, and when. 
Our dataset is similar in spirit to the TRECVID dataset but ours is larger. 
TRECVID contains about 2k videos with 10k captions, while our dataset has approximately 80k videos with 399k captions.

\section{Sentence Generation}
\label{sec:generation}

\begin{figure}[tb]
  \small
  \centering
  \includegraphics[scale=0.24]{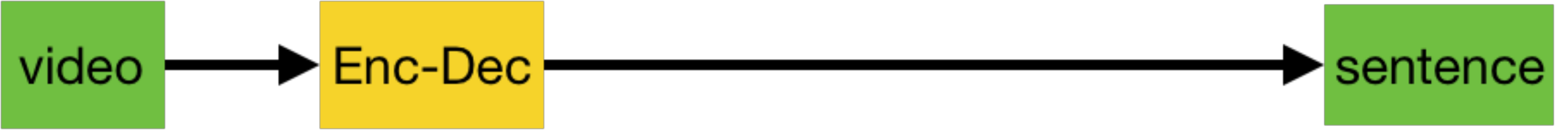} \\
  (a) Sentence-wise approach \\[10pt]
  \includegraphics[scale=0.24]{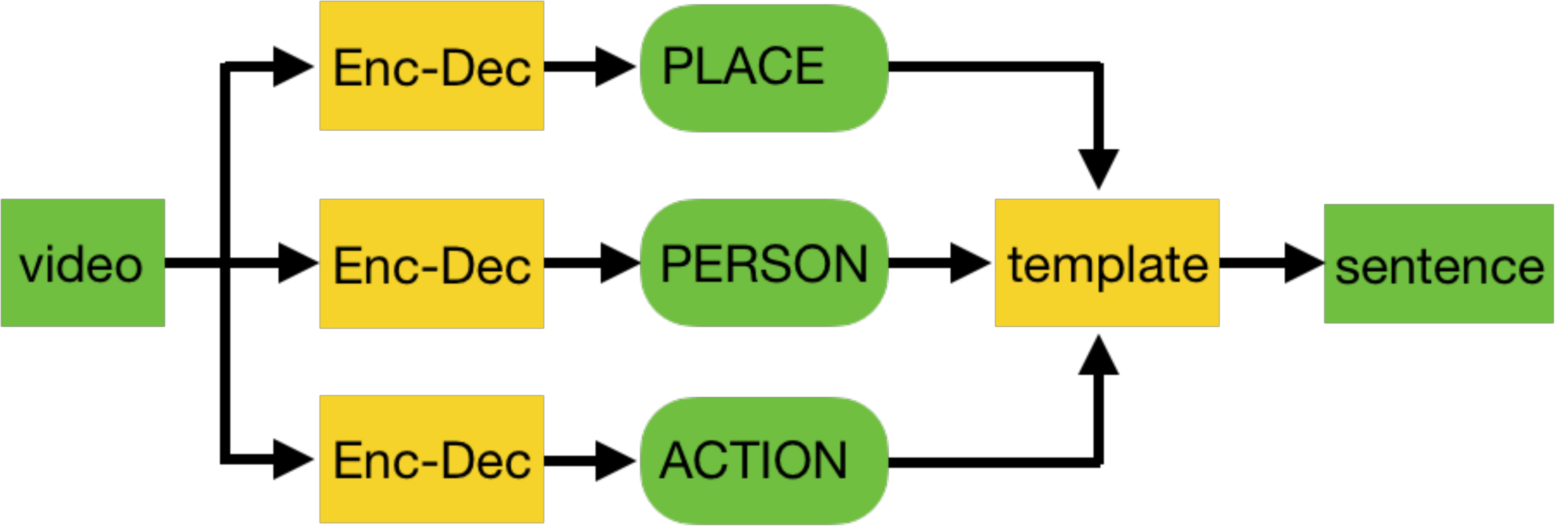} \\
  (b) Phrase-wise approach
  \caption{
    Overview of two sentence generation approaches.
    (a) A sentence-wise generation approach generates a sentence directory using a single encoder-decoder model.
    (b) A phrase-wise approach first generates three phrases (PLACE, PERSON, and ACTION) separately, 
    and then outputs a complete sentence using the template.
  }
  \label{fig:method}
\end{figure}

We evaluated two sentence generation approaches: sentence-wise and phrase-wise approaches. 
Figure~\ref{fig:method} presents an overview of both. 

The sentence-wise approach generates a whole sentence using a single encoder-decoder model; a standard approach in the video captioning literature. 

The phrase-wise approach uses three encoder-decoders. 
It first generates PLACE, PERSON, and ACTION, respectively, 
and then fills in slots in the template (i.e., PLACE で PERSON が ACTION). 
This approach is a reasonable way of achieving the current objective of generating a description that specifies ``who does what and where.'' 
Another advantage to this approach is that the training decoders for phrase generation is easier than for sentence generation. 
Since phrases are shorter than sentences, 
it is sufficient for the decoders to target relatively short sequences of words.

In our experiments, 
we used a multi-modality fusion caption generation method~\cite{jin2016describing}---the winning solution of the MSR Video to Language Challenge 2016---as the encoder-decoder model in both sentence-wise and phrase-wise approaches; a method frequently used as a baseline in video captioning experiments. 
In this method, 
the encoder (a multilayer feedforward network) transforms multi-modality features into a single vector, 
and the decoder (a recurrent neural network) generates a sequence of words from the vector. 

The output word sequence is chosen by beam search. 
To eliminate length bias,
we used the length normalization presented in \newcite{wu2016google}.

\section{Experiment}
\label{sec:experiment}

\begin{table*}[t]
  \small
  \centering

  \begin{tabular}{l rrr rrr rrr}
    \toprule
    & \multicolumn{3}{c}{PLACE} & \multicolumn{3}{c}{PERSON} & \multicolumn{3}{c}{ACTION} \\
    \cmidrule(lr){2-4} \cmidrule(lr){5-7} \cmidrule(lr){8-10}
    modality                    & BLEU        & ROUGE          & CIDEr       
                             & BLEU        & ROUGE          & CIDEr       
                             & BLEU        & ROUGE          & CIDEr       \\
    \midrule
      I + M & 0.792 & 0.855 & \textbf{1.848} & \textbf{0.732} & \textbf{0.789} & \textbf{1.725} & \textbf{0.801} & \textbf{0.866} & \textbf{3.346}  \\
      I      & \textbf{0.833} & \textbf{0.868} & 1.821 & 0.717 & 0.779 & 1.686 & 0.780 & 0.851 & 3.156  \\
      M      & 0.773 & 0.830 & 1.736 & 0.646 & 0.736 & 1.443 & 0.769 & 0.844 & 3.097  \\
    \bottomrule
  \end{tabular}

  \caption{
  Results from the phrase generation task.
  Bold figures indicate the best performer for each evaluation criterion.
  ``I'' represents the image modality 
  and ``M'' is the motion modality.
  }
  \label{tab:result-template}

\end{table*}

\begin{table}[tb]
  \small
  \centering

  \begin{tabular}{ll rrr}
    \toprule
    approach          & modality                    & BLEU & ROUGE & CIDEr \\
    \midrule
    sentence-wise        
          & I + M        & 0.713 & \textbf{0.795} & 1.837  \\
          & I            & 0.696 & 0.786 & 1.769  \\
          & M            & 0.666 & 0.769 & 1.677  \\
    \midrule
    phrase-wise
          & I + M        & \textbf{0.749} & 0.791 & \textbf{1.937}  \\
          & I            & 0.735 & 0.785 & 1.846 \\
          & M            & 0.696 & 0.765 & 1.729  \\
    \bottomrule
  \end{tabular}

  \caption{
    Results from the sentence generation task.
  }
  \label{tab:result-sentence}

\end{table}

We used sentence and phrase generation tasks to evaluate our dataset. 
The objective of this experiment was to investigate two points: 
(i) whether the methods can specify ``who does what and where'' 
and (ii) the differences between sentence-wise and phrase-wise approaches.

\subsection{Experimental Setups}
\label{sec:experiment-setup}

\paragraph{Dataset}
We randomly split the videos into training (80\%), development (10\%), and test (10\%) sets.

We ran SentencePiece\footnote{{https://github.com/google/sentencepiece}}~\cite{kudo2018}, 
an unsupervised text tokenizer, 
to segment captions into subwords.
We trained the SentencePiece model on a subset of the entire set of captions used to train the captioning methods. 
The vocabulary size of this model was set to 8,000.

\paragraph{Evaluation Criteria}
In accordance with the literature~\cite{Xiang2018video,pan2017video,gan2017semantic,wang2018,Phan:2017:MMA},
generated captions were evaluated based on three criteria; 
BLEU-4~\cite{papineni2002bleu}, ROUGE-L~\cite{lin2004rouge}, and CIDEr~\cite{vedantam2015cider}.
In the evaluation phase, 
we first tokenized the generated captions and references using KyTea,
and then computed scores.

\paragraph{Hyperparameters}
We used a gated recurrent unit as recurrent neural network (RNN) cell, 
and tuned the following hyperparameters: 
RNN hidden state size, 
RNN layer size, 
learning rate, 
weight decay, dropout probability, beam width, 
and length normalization coefficient. 
We chose those with the best CIDEr score on the development set. 

\paragraph{Input Modality}
We used image and motion modalities as encoder inputs. 
The image modality captures static image content from video frames. 
In accordance with previous work \cite{Xiang2018video,wang2018},
we used the last layer of the ResNet-152~\cite{he2016deep} trained on ImageNet.\footnote{
{https://pytorch.org/docs/master/torchvision/models.html}}
First, we sampled frames at 3~fps 
and then extracted a 2,048 dimensional vector from each frame. 
The motion modality captures the local temporal motion. 
We used 3D ResNeXt-101~\cite{hara3dcnns} trained on Kinetics-400.\footnote{
{https://github.com/kenshohara/video-classification-3d-cnn-pytorch}}
We first split a video into a set of 16 frames 
and then converted each set (16 frames) to a 2,048-dimensional vector. 
In both modalities, we used mean pooling to aggregate the vectors obtained from a video.

\subsection{Experimental Results}
\label{sec:experiment-result}

Table~\ref{tab:result-template} shows results from the phrase generation task.
In all methods except PLACE,
the best results were obtained when two modalities were input (I + M). 
In PLACE, 
use of the image modality alone was found to be more efficient. 
This is as expected because generating a phrase for PLACE does not require information about local temporal motion. 
Consequently, the generator with two modalities did not affect the results. 

Table~\ref{tab:result-sentence} shows the results of sentence generation.
The use of two modalities with both sentence-wise and phrase-wise generation performed better than the single modality cases across all criteria, 
and image modality alone came second.

We found the phrase-wise approach outperformed sentence-wise generations in BLEU and CIDEr. 
In ROUGE, the sentence-wise approach was observed to be slightly better than the phrase-wise approach.

\subsection{Generated Captions}

\begin{figure*}[!htbp]
  \small
  \centering

  \includegraphics[scale=0.125]{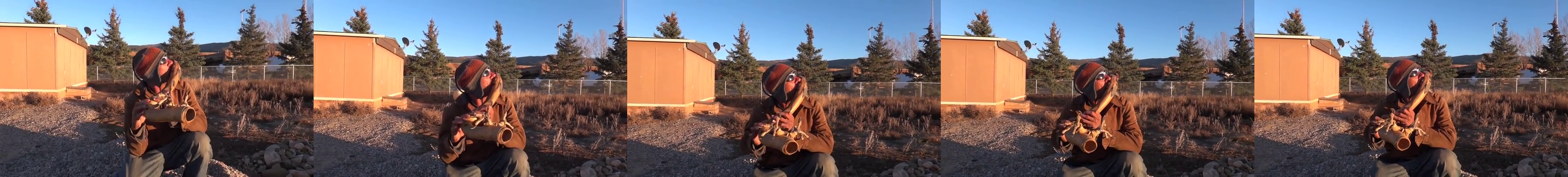} \\[5pt]
  \begin{tabular}{ll}
    \multicolumn{2}{c}{(a) Input video (1 fps).} \\[5pt]
    \toprule
    method & description \\
    \midrule
    sentence-wise & 木が生えている屋外で茶色い服を着た男性がカメラで写真を撮っている\\
    &(A man worn brown clothes is taking a photo by using a camera in the open air with woody) \\[5pt]
    phrase-wise   & 屋外で茶色い服の男性が笛を吹いている \\
    & (A man with brown clothes is blowing a horn in the open air) \\[5pt]
    \cmidrule(lr){0-1}
    Human annotation
    & 屋外でサングラスをした男性が楽器を演奏している \\
    & (A man worn sunglasses in the open air is playing an instrument) \\[5pt]
    & 外で茶色の服を着ている男性が笛を吹いている\\
    & (A man is wearing brown clothes who is blowing a horn in the open air) \\[5pt]
    & 屋外で帽子にサングラスをした男性が楽器を演奏している\\
    & (A man worn a hat and sunglasses is playing an instrument in the open air) \\[5pt]
    & 屋外で帽子とサングラスをした人がしゃがんで角笛を吹いている\\
    & (A person worn a hat and sunglasses in the open air is squatting eyes and blowing a horn) \\[5pt]
    & フェンスがある屋外で帽子をかぶってサングラスをかけた男性が楽器を演奏している \\
    & (A man worn a hat and sunglasses is playing an instrument in the open air with a fence) \\[5pt]
    \bottomrule
    \multicolumn{2}{c}{(b) Reference descriptions and generated sentences.}
  \end{tabular}
  \caption{
    An example of ground truth descriptions and sentences generated by the sentence-wise and phrase-wise methods.  }
  \label{fig:result-phrase}
\end{figure*}

\begin{figure*}[!htbp]
  \small
  \centering

  \includegraphics[scale=0.125]{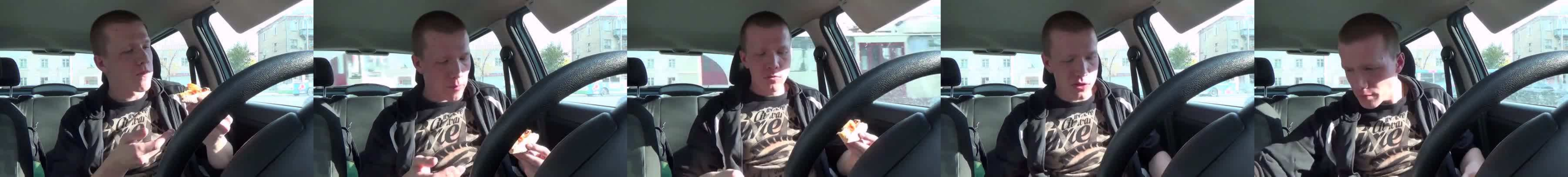} \\[5pt]
  \begin{tabular}{ll}
    \multicolumn{2}{c}{(a) Input video (1 fps).} \\[10pt]
    \toprule
    method & description \\
    \midrule
    sentence-wise & 車内で坊主頭の男性が食事をしている\\
    &(A man with a shaven head is eating in a car) \\[5pt]
    phrase-wise   & 車内で黒い服を着た男性が話している \\
    & (A man worn black clothes is speaking in a car) \\[5pt]
    \cmidrule(lr){0-1}
    Human annotation
    & 車内で坊主頭の男性がピザを食べている \\
    & (A man with a shaven head is eating pizza in a car) \\[5pt]
    & 車内で黒い服の男性が何かを食べている\\
    & (A man with black clothes is eating something in a car) \\[5pt]
    & 車の中で黒い服を着た男性が何かを食べている \\
    & (A man worn black clothes is eating something in a car) \\[5pt]
    & 車内で黒い服を着た男性が食べ物を食べている \\
    & (A man worn black clothes is eating something in a car) \\[5pt]
    & 車内で上着を着た短髪の男性が運転席に座った状態で食べ物を食べている \\
    & (A short-haired man worn jacket is eating food on the driver's seat in a car) \\[5pt]
    \bottomrule
    \multicolumn{2}{c}{(b) Reference descriptions and generated sentences.}
  \end{tabular}
  \caption{
    An example of ground truth descriptions and sentences generated by the sentence-wise and phrase-wise methods.
  }
  \label{fig:result-sentence}
\end{figure*}

\begin{figure*}[!th]
  \small
  \centering
  
  \includegraphics[scale=0.125]{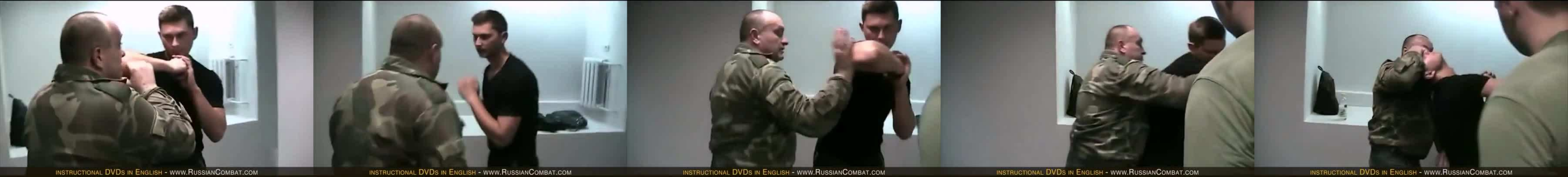} \\
  \begin{tabular}{ll}
    \multicolumn{2}{c}{(c) Input video (1 fps).} \\[5pt]
    \toprule
    method & description \\
    \midrule
    sentence-wise & 白い壁の部屋の中で2人の男性が抱き合っている \\
    & (Two men are hugging in a room with white wall) \\[5pt]
    phrase-wise   & 白い壁の部屋で迷彩服の男性が抱き合っている \\
    & (Men with camouflage clothes are hugging in a room with white wall) \\[5pt]
    \cmidrule(lr){0-1}
    Human annotation 
    & 屋内で迷彩服の男性が格闘術を教えている\\
    & (A man with camouflage clothes is teaching hand-to-hand combat inside the room) \\[5pt]
    & 屋内で黒いティーシャツの男性がおさえこまれている \\
    & (A man with a black T-shirt is being arrested inside the room) \\[5pt]
    & 白い壁の部屋で黒い服を着た男性がサバイバルの訓練をしている \\
    & (A man worn black clothes is training survival skills in a room with white wall) \\[5pt]
    & 白い壁の部屋の中で黒い服を着た男性が緑色の服を着た男性を殴っている \\
    & (A man worn black clothes is hitting a man worn green clothes in a room with white walls) \\[5pt]
    & 白い壁で薄暗い屋内でカーキや黒のトップスを着た体格のいい男性たちが護身術を教わっている \\
    & (Muscular men worn khaki and black clothes are being taught self-defense in a dim room with white wall) \\[5pt]
    \bottomrule
    \multicolumn{2}{c}{(d) Reference descriptions and generated sentences.}
  \end{tabular}

  \caption{
    An example of ground truth descriptions and sentences generated by the sentence-wise and phrase-wise methods.
  }
  \label{fig:result-both}
\end{figure*}

We presented three samples of generated captions and references.
Figure~\ref{fig:result-phrase} shows that the phrase-wise approach captured the action (\textit{blowing a horn}) of the input video, while the sentence-wise approach generated the wrong action phrase (\textit{taking a photo}).
Contrary to these results, the captions generated by the sentence-wise approach, shown in figure~\ref{fig:result-sentence}, are better than those of the phrase-wise approach. 
In Figure~\ref{fig:result-both}, neither approach generated accurate action phrases.

\section{Conclusion}
\label{sec:conclusion}

We constructed a new video caption dataset for describing human actions in Japanese. 
The advantage of this dataset is that the captions are written in Japanese and specify ``who does what and where.''
To specify this, we conducted two procedures: Phrase annotation and template-based sentence construction. 
Although the template-based construction does not produce grammatically varied sentences, the sentences produced are not unnatural. 
Our dataset, consisting of 79,822 videos and 399,233 captions, 
is the first Japanese caption dataset, 
and the largest video caption dataset in any language with respect to the number of captions. 

We evaluated two approaches based on a multi-modality fusion caption generation method on our dataset: Sentence-wise and phrase-wise approaches. 
Experiments showed that the phrase-wise approach outperformed the sentence-wise approach with respect to BLEU and CIDEr. 
In addition, we evaluated phrase generation quality using our dataset, 
employing phrase generation tasks to ascertain whether the generation methods specified ``who does what and where.'' 
We observed that the image and motion modalities to be useful in explaining PERSON and ACTION, while image modality alone was sufficient for PLACE.

\newpage

\clearpage

\section{Acknowledgements}

We thank anonymous reviewers for their valuable comments and suggestions. 
This paper is based on results obtained from a project commissioned by the New Energy and Industrial Technology Development Organization (NEDO).

\section{Bibliographical References}

\bibliography{caption,image-caption,tool,video-caption}

\begin{thebibliography}{}

\bibitem[\protect\citename{Awad \bgroup et al.\egroup }2018]{2018trecvidawad}
Awad, G., Butt, A., Curtis, K., Lee, Y., Fiscus, J., Godil, A., Joy, D.,
  Delgado, A., Smeaton, A.~F., Graham, Y., Kraaij, W., Quénot, G., Magalhaes,
  J., Semedo, D., and Blasi, S.
\newblock (2018).
\newblock {TRECVID 2018: Benchmarking Video Activity Detection, Video
  Captioning and Matching, Video Storytelling Linking and Video Search}.
\newblock In {\em TRECVID}.

\bibitem[\protect\citename{Chen and Dolan}2011]{Chen2010}
Chen, D.~L. and Dolan, W.~B.
\newblock (2011).
\newblock {Collecting Highly Parallel Data for Paraphrase Evaluation}.
\newblock In {\em ACL}, pages 190--200.

\bibitem[\protect\citename{Gan \bgroup et al.\egroup }2017]{gan2017semantic}
Gan, Z., Gan, C., He, X., Pu, Y., Tran, K., Gao, J., Carin, L., and Deng, L.
\newblock (2017).
\newblock {Semantic Compositional Networks for Visual Captioning}.
\newblock In {\em CVPR}, pages 5630--5639.

\bibitem[\protect\citename{Gella \bgroup et al.\egroup }2018]{D18-1117}
Gella, S., Lewis, M., and Rohrbach, M.
\newblock (2018).
\newblock {A Dataset for Telling the Stories of Social Media Videos}.
\newblock In {\em EMNLP}, pages 968--974.

\bibitem[\protect\citename{Hara \bgroup et al.\egroup }2018]{hara3dcnns}
Hara, K., Kataoka, H., and Satoh, Y.
\newblock (2018).
\newblock {Can Spatiotemporal 3D CNNs Retrace the History of 2D CNNs and
  ImageNet?}
\newblock In {\em CVPR}, pages 6546--6555.

\bibitem[\protect\citename{He \bgroup et al.\egroup }2016]{he2016deep}
He, K., Zhang, X., Ren, S., and Sun, J.
\newblock (2016).
\newblock {Deep Residual Learning for Image Recognition}.
\newblock In {\em CVPR}, pages 770--778.

\bibitem[\protect\citename{Jin \bgroup et al.\egroup }2016]{jin2016describing}
Jin, Q., Chen, J., Chen, S., Xiong, Y., and Hauptmann, A.
\newblock (2016).
\newblock {Describing Videos Using Multi-Modal Fusion}.
\newblock In {\em ACM MM}, pages 1087--1091.

\bibitem[\protect\citename{Krishna \bgroup et al.\egroup }2017]{Krishna2017a}
Krishna, R., Hata, K., Ren, F., Fei-Fei, L., and Niebles, J.~C.
\newblock (2017).
\newblock {Dense-Captioning Events in Videos}.
\newblock In {\em ICCV}, pages 706--715.

\bibitem[\protect\citename{Kudo and Richardson}2018]{kudo2018}
Kudo, T. and Richardson, J.
\newblock (2018).
\newblock {SentencePiece: A Simple and Language Independent Subword Tokenizer
  and Detokenizer for Neural Text Processing}.
\newblock In {\em EMNLP: System Demonstrations}, pages 66--71.

\bibitem[\protect\citename{Lin}2004]{lin2004rouge}
Lin, C.-Y.
\newblock (2004).
\newblock {ROUGE: A Package for Automatic Evaluation of Summaries}.
\newblock {\em Text Summarization Branches Out}.

\bibitem[\protect\citename{Long \bgroup et al.\egroup }2018]{Xiang2018video}
Long, X., Gan, C., and de~Melo, G.
\newblock (2018).
\newblock {Video Captioning with Multi-Faceted Attention}.
\newblock {\em TACL}, 6:173--184.

\bibitem[\protect\citename{Neubig \bgroup et al.\egroup
  }2011]{neubig2011pointwise}
Neubig, G., Nakata, Y., and Mori, S.
\newblock (2011).
\newblock {Pointwise Prediction for Robust, Adaptable Japanese Morphological
  Analysis}.
\newblock In {\em ACL}, pages 529--533.

\bibitem[\protect\citename{Pan \bgroup et al.\egroup }2017]{pan2017video}
Pan, Y., Yao, T., Li, H., and Mei, T.
\newblock (2017).
\newblock {Video Captioning with Transferred Semantic Attributes}.
\newblock In {\em CVPR}, pages 984--992.

\bibitem[\protect\citename{Papineni \bgroup et al.\egroup
  }2002]{papineni2002bleu}
Papineni, K., Roukos, S., Ward, T., and Zhu, W.-J.
\newblock (2002).
\newblock {BLEU: A Method for Automatic Evaluation of Machine Translation}.
\newblock In {\em ACL}, pages 311--318.

\bibitem[\protect\citename{Phan \bgroup et al.\egroup }2017]{Phan:2017:MMA}
Phan, S., Miyao, Y., and Satoh, S.
\newblock (2017).
\newblock {MANet: A Modal Attention Network for Describing Videos}.
\newblock In {\em Proceedings of the 2017 ACM on Multimedia Conference}, pages
  1889--1894.

\bibitem[\protect\citename{Rohrbach \bgroup et al.\egroup }2014]{101007}
Rohrbach, A., Rohrbach, M., Qiu, W., Friedrich, A., Pinkal, M., and Schiele, B.
\newblock (2014).
\newblock {Coherent Multi-sentence Video Description with Variable Level of
  Detail}.
\newblock In {\em German Conference on Pattern Recognition}, pages 184--195.

\bibitem[\protect\citename{Rohrbach \bgroup et al.\egroup }2017]{Rohrbach2017}
Rohrbach, A., Torabi, A., Rohrbach, M., Tandon, N., Pal, C., Larochelle, H.,
  Courville, A., and Schiele, B.
\newblock (2017).
\newblock {Movie Description}.
\newblock {\em IJCV}, 123(1):94--120.

\bibitem[\protect\citename{Sigurdsson \bgroup et al.\egroup
  }2016]{Sigurdsson2016}
Sigurdsson, G.~A., Varol, G., Wang, X., Farhadi, A., Laptev, I., and Gupta, A.
\newblock (2016).
\newblock {Hollywood in Homes: Crowdsourcing Data Collection for Activity
  Understanding}.
\newblock In {\em ECCV}, pages 510--526.

\bibitem[\protect\citename{Vedantam \bgroup et al.\egroup
  }2015]{vedantam2015cider}
Vedantam, R., Lawrence~Zitnick, C., and Parikh, D.
\newblock (2015).
\newblock {CIDEr: Consensus-Based Image Description Evaluation}.
\newblock In {\em CVPR}, pages 4566--4575.

\bibitem[\protect\citename{Venugopalan \bgroup et al.\egroup
  }2015a]{venugopalan2015sequence}
Venugopalan, S., Rohrbach, M., Donahue, J., Mooney, R., Darrell, T., and
  Saenko, K.
\newblock (2015a).
\newblock {Sequence to Sequence-Video to Text}.
\newblock In {\em ICCV}, pages 4534--4542.

\bibitem[\protect\citename{Venugopalan \bgroup et al.\egroup
  }2015b]{venugopalan2014translating}
Venugopalan, S., Xu, H., Donahue, J., Rohrbach, M., Mooney, R., and Saenko, K.
\newblock (2015b).
\newblock {Translating Videos to Natural Language Using Deep Recurrent Neural
  Networks}.
\newblock In {\em NAACL}, pages 1494--1504.

\bibitem[\protect\citename{Wang \bgroup et al.\egroup }2018]{wang2018}
Wang, X., Wang, Y.-F., and Wang, W.~Y.
\newblock (2018).
\newblock {Watch, Listen, and Describe: Globally and Locally Aligned
  Cross-Modal Attentions for Video Captioning}.
\newblock In {\em NAACL}, pages 795--801.

\bibitem[\protect\citename{Wu \bgroup et al.\egroup }2016]{wu2016google}
Wu, Y., Schuster, M., Chen, Z., Le, Q.~V., Norouzi, M., Macherey, W., Krikun,
  M., Cao, Y., Gao, Q., Macherey, K., et~al.
\newblock (2016).
\newblock {Google's Neural Machine Translation System: Bridging the Gap Between
  Human and Machine Translation}.
\newblock {\em arXiv preprint arXiv:1609.08144}.

\bibitem[\protect\citename{Xu \bgroup et al.\egroup }2016]{Xu2016}
Xu, J., Mei, T., Yao, T., and Rui, Y.
\newblock (2016).
\newblock {MSR-VTT: A Large Video Description Dataset for Bridging Video and
  Language}.
\newblock In {\em CVPR}, pages 5288--5296.

\bibitem[\protect\citename{Yao \bgroup et al.\egroup }2015]{yao2015describing}
Yao, L., Torabi, A., Cho, K., Ballas, N., Pal, C., Larochelle, H., and
  Courville, A.
\newblock (2015).
\newblock {Describing Videos by Exploiting Temporal Structure}.
\newblock In {\em ICCV}, pages 4507--4515.

\bibitem[\protect\citename{Yoshikawa \bgroup et al.\egroup
  }2018]{stairactions2018}
Yoshikawa, Y., Lin, J., and Takeuchi, A.
\newblock (2018).
\newblock {STAIR Actions: A Video Dataset of Everyday Home Actions}.
\newblock {\em arXiv preprint arXiv:1804.04326}.

\bibitem[\protect\citename{Zhou \bgroup et al.\egroup }2018]{ZhXuCoCVPR18}
Zhou, L., Xu, C., and Corso, J.~J.
\newblock (2018).
\newblock {Towards Automatic Learning of Procedures From Web Instructional
  Videos}.
\newblock In {\em AAAI}, pages 7590--7598.

\end{thebibliography}
\bibliographystyle{lrec}

\end{document}